\documentclass{article}
\usepackage{ijcai26}

\usepackage{times}
\usepackage{microtype}
\usepackage{soul}
\usepackage{url}
\usepackage{natbib} 
\usepackage[hidelinks]{hyperref}
\usepackage[utf8]{inputenc}
\usepackage[small]{caption}
\usepackage{subcaption}
\usepackage{xcolor}
\usepackage{graphicx}
\usepackage{amsmath}
\usepackage{amssymb}
\usepackage{mathtools}
\usepackage{amsthm}
\usepackage{empheq}
\usepackage{booktabs}
\usepackage{algorithm}
\usepackage{algorithmic}

\title{When Context Compensates for Sparse Event History:\\
AlphaEarth for Spatio-Temporal Point-Process Forecasting}

\author{
Yahya Aalaila$^{1,5}$\and
Mouad Elhamdi$^2$\and
Gerrit Gro{\ss}mann$^1$\and
Daniel Jenson$^3$\and
Elizaveta Semenova$^4$\And
Sebastian Vollmer$^{1,5}$\\
\affiliations
$^1$German Research Center for Artificial Intelligence (DFKI)\\
$^2$Universit\'e Mohammed VI Polytechnique\\
$^3$University of Oxford\\
$^4$Imperial College London\\
$^5$Rhineland-Palatinate Technical University of Kaiserslautern-Landau (RPTU) \\
\emails
yahya.aalaila@dfki.de
}

\begin{document}

\maketitle
\begin{abstract}
Spatio-temporal point-process models must often generalise across space when local event histories are sparse. We study whether exogenous spatial context can compensate in such regimes. Using a fixed log-Gaussian Cox process backbone, we compare an event-only model with the same model augmented by AlphaEarth embeddings as linear spatial context. We evaluate spatial transfer on emergency medical services (EMS) forecasting across eight held-out regions, fixed forecast anchors, and a sweep over history length $w$, using only AlphaEarth (AE) embeddings available strictly before each anchor. AE improves out-of-region predictive performance across all history regimes, with the largest gains under scarce histories: approximately $2$--$6\times$ multiplicative improvements at $1-2$ weeks, tapering to roughly $10$--$20\%$ at $w=20$--$104$ weeks. These results show that contextual information can substantially stabilise spatially transferred point-process forecasts when event history is limited.
\end{abstract}

\section{INTRODUCTION}\label{sec:introduction}

Spatio-temporal point-process (STPP) models forecast future events by estimating how risk varies over space and time from previously observed event histories. Yet they are often asked to generalize precisely where that history is weakest. Consider an emergency medical services (EMS) agency forecasting next-week call demand in a newly developed tract, a boundary region with limited prior coverage, or a neighborhood whose recent call history is too sparse to reveal a stable risk pattern. A history-driven point-process model may observe only a handful of local events, even though the area already has observable spatial structure: its built environment, land-use pattern, surrounding settlement density, and broader geospatial context. This raises a basic question: \emph{when event history is scarce, can external spatial context compensate?}

STPPs are a natural setting for this question because they forecast future events from observed histories while allowing exogenous covariates to shape the conditional intensity \citep{diggle2013statistical,diggle2013spatial,reinhart2018review}. This exposes an endogenous--exogenous trade-off: event histories reflect where the process has recently concentrated, while contextual information describes the structure of the domain in which those events occur. We lack controlled evidence on whether such context can stabilise spatial transfer when local history is scarce, and how its value changes as more event history becomes available.
\begin{figure}[t]
    \centering
    \includegraphics[width=\linewidth]{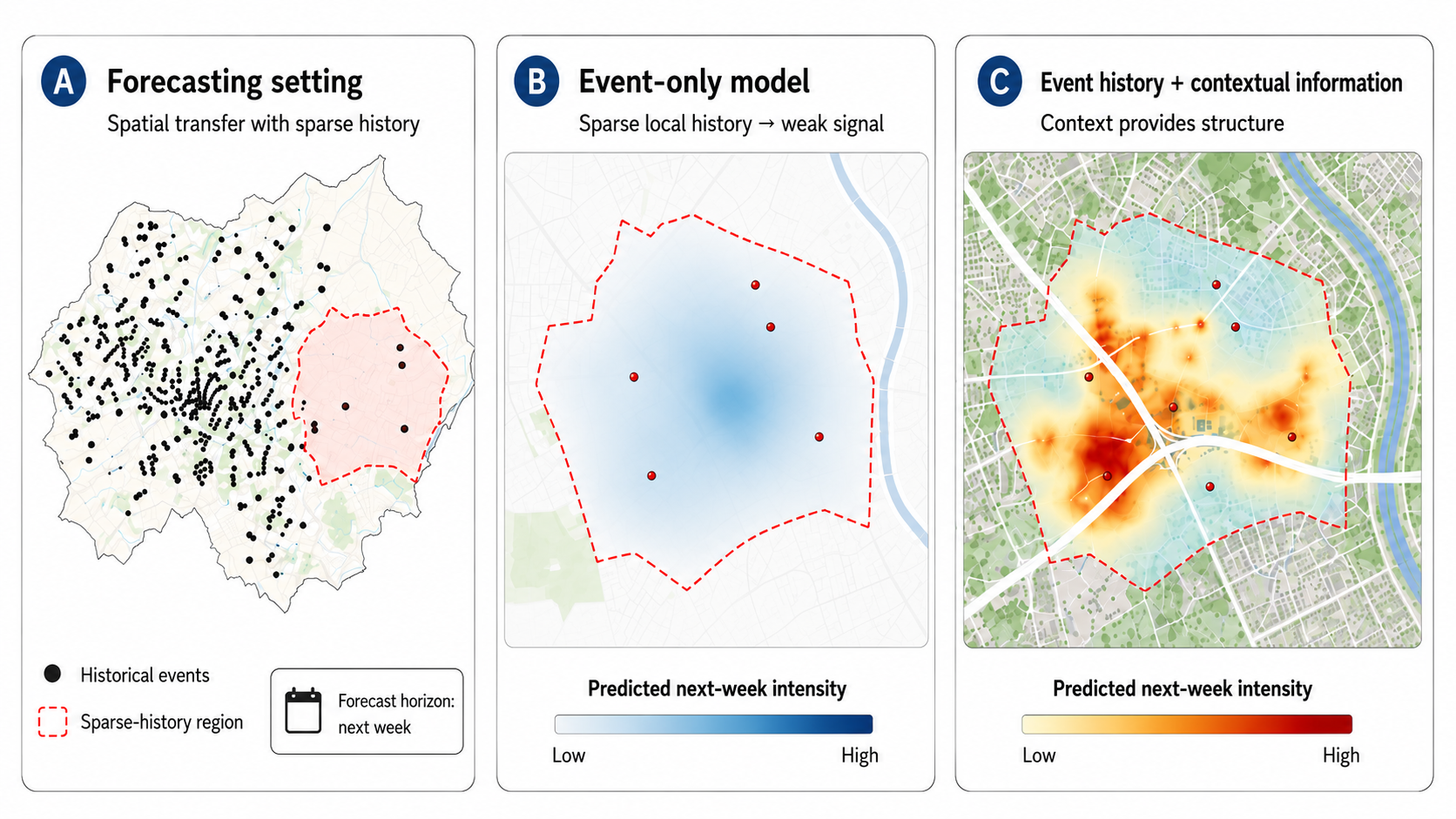}
\caption{
Conceptual motivation. In spatial transfer, contextual information can stabilise point-process forecasts when local event history is sparse.
}
    \label{fig:history_context_motivation}
\end{figure}
This trade-off matters because spatial event risk is shaped by persistent structure in the domain---built environment, mobility networks, land cover, infrastructure, and other properties of place---that event histories can only recover indirectly, especially under short observation windows or spatial transfer. Studying the role of such context has historically required bespoke remote-sensing pipelines and hand-engineered spatial features \citep{claverie2018harmonized}. AlphaEarth (AE) embeddings make the question easier to examine cleanly: they provide standardized, foundation-model geospatial representations that compress multi-source satellite and environmental signals into compact 64-dimensional vectors per location and year \citep{brown2025alphaearth}. We use AE embeddings as a controlled source of exogenous spatial context to test how contextual information complements event history across history regimes. Figure~\ref{fig:history_context_motivation} illustrates this setting.
{
Our goal is not to show that exogenous covariates can help STPPs in general, which is already well established, but to quantify when off-the-shelf foundation-model geospatial embeddings improve spatial transfer relative to accumulating local event history.
}

We begin from a simple intuition: external spatial context should matter most when local event history is scarce, and its incremental value should diminish as the event-only model accumulates enough data to recover stable background structure. We test this intuition in a controlled EMS forecasting study. To avoid conflating contextual information with model capacity, we fix a transparent LGCP backbone and compare two matched models: an event-only LGCP and the same LGCP augmented with AE embeddings as linear spatial context. 
We evaluate geographically held-out forecasting across eight disjoint spatial masks within Montgomery County, Pennsylvania, under fixed forecast anchors and a sweep over history-window lengths $w$. AE is treated as strictly exogenous: for each forecast anchor, we use only the latest annual embedding slice available before the anchor and hold it fixed during training and evaluation.

The results support the history--context intuition strongly. Across all held-out regions, AE improves out-of-region predictive performance for every history length considered. The gains are largest in sparse-history regimes: with only 1-2 weeks of training data, AE yields approximately $2$--$6\times$ multiplicative improvements in predictive density relative to the event-only baseline. The advantage narrows as more history accumulates, but remains positive even for $20$--$104$ weeks of training history, where gains persist at roughly $10$--$20\%$. Posterior field summaries further indicate that AE accelerates early level correction and produces smoother, more stable spatial structure.

\section{PROBLEM SETUP}
\label{sec:problemsetup}
We study LGCPs for spatio-temporal events under varying history-window regimes, with and without exogenous spatial context. The goal is to quantify how history length $w$ and AE embeddings affect spatially held-out predictive performance.
\subsection{Models and Training}
\label{subsec:models}
In an LGCP, the conditional intensity at $(t,s)$ does not explicitly condition on the event history at prediction time; rather, the chosen history window determines which past events enter estimation. The model assumes,
\begin{empheq}[left=\empheqlbrace]{align*}
\lambda(t,\boldsymbol{s}) &= \exp\bigl\{\eta(t,\boldsymbol{s})\bigr\}, \\
\eta(t,\boldsymbol{s}) &= \alpha_{\beta}(t,\boldsymbol{s}) + g(t,\boldsymbol{s}),
\end{empheq}
where $g \sim \mathrm{GP}(0,k)$ captures smooth spatio-temporal variation, and $\alpha_\beta(t,\boldsymbol{s})$ collects fixed effects and contextual terms.

For computational convenience, we model the spatiotemporal effect as an additive term, $g(t,s)=f_s(s)+f_t(t)$. This decomposition reduces inference cost while still allowing separate spatial and temporal variation. We also exclude neural components because they add flexibility that can improve fit independently of the added spatial context. By keeping the model simple, observed performance differences can be attributed to the added spatial context rather than to increased model complexity. In this study, we consider two models:
\paragraph{Event-only LGCP.}
The intensity is
\begin{equation}
\label{eq:baseline}
\lambda_{\mathrm{E}}(t,\boldsymbol{s})
=
\exp\{\beta_0 + f_s(\boldsymbol{s}) + f_t(t)\}.
\end{equation}
\paragraph{AE-augmented LGCP.}
We augment the log-intensity with AE spatial context,

\begin{equation}
    \label{eq:baseline+AE}
    \lambda_{\mathrm{AE}}(t,\boldsymbol{s})
    =
    \exp\{\beta_0 + \boldsymbol{z}(t,\boldsymbol{s})^\top\beta
    + f_s(\boldsymbol{s}) + f_t(t)\}.
\end{equation}


AE vectors $\boldsymbol{z}(t,\boldsymbol{s})\in\mathbb{R}^{64}$ are piecewise-constant in time at an annual cadence: for $t\in(\tau_y,\tau_{y+1}]$, $\boldsymbol{z}(t,\boldsymbol{s})=\boldsymbol{z}_y(\boldsymbol{s})$. For a forecasting anchor $t_a$, we use the last annual slice strictly preceding the anchor and hold it fixed during training and evaluation.

\paragraph{Forecasting protocol.}
We partition \(\mathcal{S}\) into disjoint regions
\(R_{\mathrm{in}}^{(m)}\) and \(R_{\mathrm{out}}^{(m)}\) for mask \(m\). Estimation uses only events in
\(R_{\mathrm{in}}^{(m)}\), while evaluation uses only
\(R_{\mathrm{out}}^{(m)}\). Anchor times \(\{t_a\}\) are chosen in
July~2020 via a back-off heuristic to ensure non-empty windows. In total, we evaluate 5 forecast anchors per mask. For anchor
\(t_a\), we use the last available annual AE slice with timestamp strictly
less than \(t_a\) and hold it fixed over the forecast horizon. The forecast
horizon is \([t_a,t_a+H)\), with \(H=7\) days, and the fitting interval is
\([t_a-w,t_a)\). We sweep
\(w\in\mathcal{W}=\{1,2,4,\ldots,104\}\) weeks.

For each mask \(m\), anchor \(t_a\), and history length \(w\), the fitting data
consist of events in \(R_{\mathrm{in}}^{(m)}\times[t_a-w,t_a)\); we denote these
events by \(\mathcal D_w\). Evaluation is performed on the held-out domain
\(\mathcal D_T = R_{\mathrm{out}}^{(m)}\times[t_a,t_a+H)\), with observed test
events \(\mathcal Y_T\subset \mathcal D_T\).

\paragraph{Estimation and inference.}
We use the PriorVAE reparameterization of the spatial GP prior \citep{semenova2022priorvae}, representing $g$ through a low-dimensional latent variable decoded back to the field. This improves inference tractability for the repeated $(w,t_a)$ sweeps. We fit all models with stochastic variational inference (SVI).

For a fitting window, the log-likelihood is
\[
\mathcal{L}(\beta,\theta,g)
=
\sum_{(t_i,\boldsymbol{s}_i)\in \mathcal Y_w^{(m,a)}}
\log \lambda(t_i,\boldsymbol{s}_i)
-
\int_{\Omega_w^{(m,a)}} \lambda(t,\boldsymbol{s})\,d\boldsymbol{s}\,dt .
\]
where $\Omega_w$ is the spatio--temporal region induced by the history window and forecast setup. All integrals are evaluated by numerical quadrature consistent with the model’s discretization.
\subsection{Evaluation}

Using the notation above, let \(\mathcal Y_w\) denote the fitting events for a
given mask \(m\), anchor \(t_a\), and history length \(w\). Evaluation is carried
out on the held-out test domain
\(\mathcal D_T=R_{\mathrm{out}}^{(m)}\times[t_a,t_a+H)\), with observed test
events \(\mathcal Y_T\subset\mathcal D_T\). For $M\in\{E,\mathrm{AE}\}$, let $p_M(\cdot\mid Y_w)$ denote the posterior predictive density under the event-only or AE-augmented model.

\paragraph{Held-out ELPD.}
We evaluate predictive performance using held-out expected log predictive
density (ELPD) on \(\mathcal D_T\) \citep{vehtari2017practical}:
\begin{equation}
\label{eq:method-ELPD}
\begin{split}
\mathrm{ELPD}_{M}(w;m,a)
&=
\sum_{y_i\in \mathcal Y_T}
\log p_M(y_i\mid\mathcal Y_w) \\
&\quad -
\mathbb{E}\!\left[
\int_{\mathcal D_T}
\lambda_M(s,t;\theta)\,ds\,dt
\;\middle|\;
\mathcal Y_w
\right].
\end{split}
\end{equation}
All integrals use the same numerical quadrature as the fitted model, including
time-bin overlaps and polygon--cell area fractions.

\paragraph{Paired contrasts and event-level density ratios.}
All comparisons are paired at the level of \((m,a,w)\), controlling for mask
geometry and anchor-specific demand. The primary contrast is
\begin{equation}
\label{eq:delta}
\Delta(w;m,a)
=
\mathrm{ELPD}_{\mathrm{AE}}(w;m,a)
-
\mathrm{ELPD}_{\mathrm{E}}(w;m,a).
\end{equation}
To isolate how much predictive density the AE model assigns to the realized
events, we also report the per-event log-density contrast
\begin{equation}
\label{eq:per-event}
\Delta_e(w;m,a)
=
\frac{1}{|\mathcal Y_T|}
\sum_{y_i\in\mathcal Y_T}
\log
\frac{
p_{\mathrm{AE}}(y_i\mid\mathcal Y_w)
}{
p_{\mathrm{E}}(y_i\mid\mathcal Y_w)
}.
\end{equation}
This quantity is measured in natural-log units per event. We additionally report
\(\exp\{\Delta_e(w;m,a)\}\), which is the geometric-mean multiplicative density
ratio assigned to realized held-out events by the AE model relative to the
baseline \citep{gneiting2007strictly}.

For each mask \(m\), we aggregate across anchors using the median. Across masks,
we summarize variation using either median and IQR ribbons or mean curves with
normal-approximate \(95\%\) confidence bands. Because the number of held-out masks is small, these bands should be read as across-mask variability summaries rather than formal population-level confidence intervals.

\paragraph{Additional summaries.}
We also report the paired percent ELPD gain,
\[
\mathrm{gain}(w;m,a)
=
\frac{\Delta(w;m,a)}
{|\mathrm{ELPD}_{\mathrm{E}}(w;m,a)|},
\]
as a scale-normalized descriptive comparison. To summarize posterior field
structure, we track the spatial standard deviation of the posterior mean
log-intensity across grid cells, with an analogous temporal summary across time
bins. These curves describe how inferred spatial and temporal structure evolves
with history length.  {The event-level density ratio $(\exp{\Delta_e})$ and the percent ELPD gain are complementary but distinct summaries: the former is a geometric-mean predictive-density ratio over realised events, while the latter normalizes the total held-out ELPD difference by the event-only ELPD.}

\section{EXPERIMENTS AND RESULTS}
\label{sec:experiments}
\subsection{Data and AE Embeddings}

\begin{figure}[t]
    \centering
    \includegraphics[width=\linewidth]{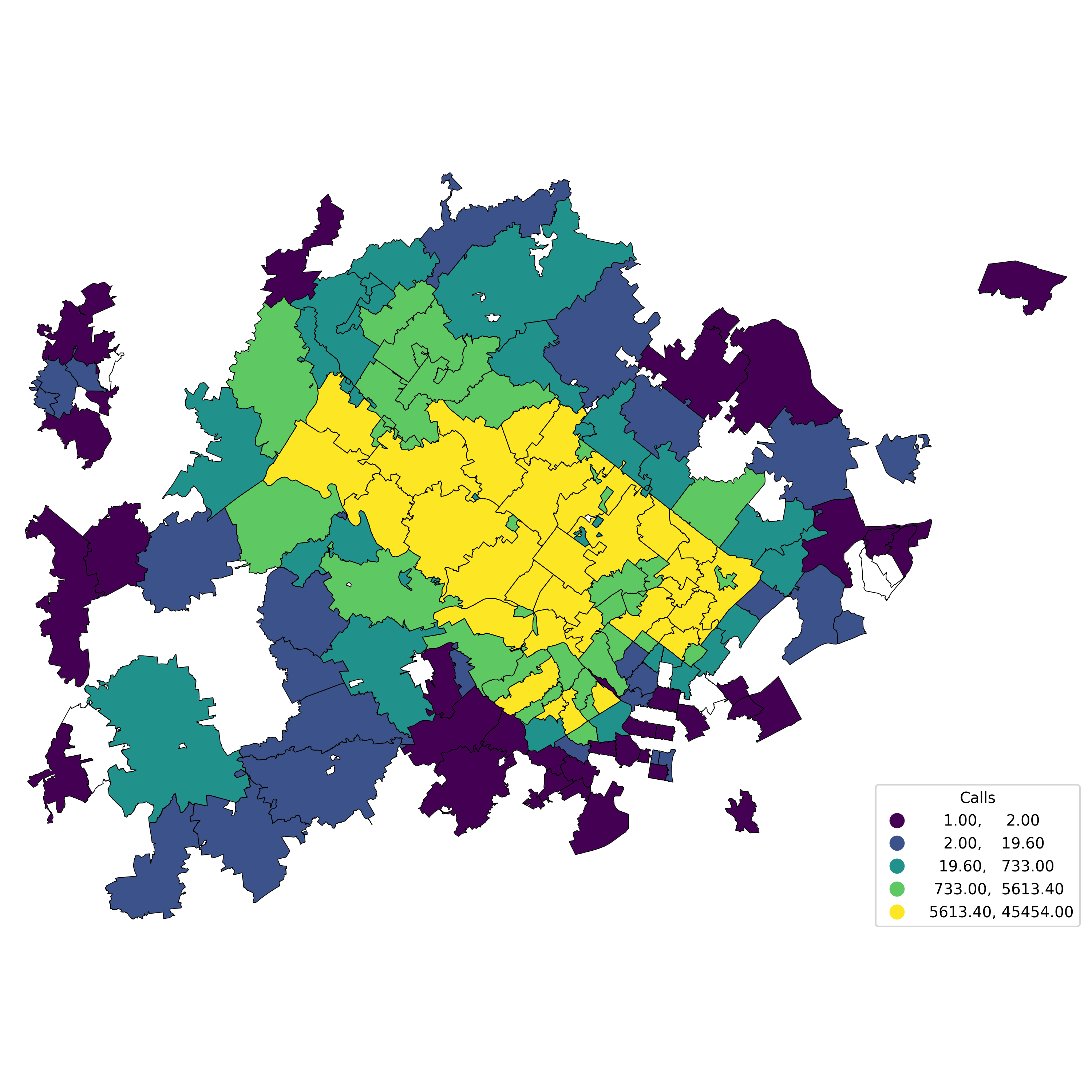}
    \caption{Spatial distribution of EMS 911 calls in Montgomery County, Pennsylvania, from 2017 to 2020.}
    \label{fig:data}
\end{figure}

We use the MontcoAlert 911 Calls dataset, focusing on EMS computer-aided dispatch incidents in Montgomery County, Pennsylvania, from 2017 to 2020. Each incident is represented as a spatio-temporal event $(t_i,\boldsymbol{s}_i)$, where $\boldsymbol{s}_i=(x_i,y_i)$ denotes planar coordinates in a metric projected coordinate reference system (CRS). The spatial domain $\mathcal{S}$ is defined by the county boundary polygon. Timestamps are converted to UTC, coordinates are reprojected to the metric CRS, and events outside $\mathcal{S}$ are removed. All spatial computations are performed in the projected CRS. Figure~\ref{fig:data} shows the spatial distribution of the resulting events.

{
Across the evaluated anchors ($40$ mask--anchor units: $8$ masks $\times$ $5$ anchors), the one-week history windows contain a mean of $1{,}908$ training events in the observed region (IQR $1{,}793$--$1{,}982$) and a mean of $262$ test events in the held-out region (IQR $179$--$381$). Training event counts increase approximately linearly with $w$ (mean slope $\approx 2{,}167$ events/week; $R^2 = 0.999$), while held-out test counts are fixed by the one-week test horizon. This clarifies that the reported history lengths correspond to genuinely sparse event regimes at small $w$.}

Following Sec.~\ref{sec:problemsetup}, we attach 64-dimensional AE embeddings
$\boldsymbol{z}(t,\boldsymbol{s})\in\mathbb{R}^{64}$ to locations in $\mathcal{S}$.
The embeddings are piecewise constant over annual intervals: for
$t\in(\tau_y,\tau_{y+1}]$, we set
$\boldsymbol{z}(t,\boldsymbol{s})=\boldsymbol{z}_y(\boldsymbol{s})$.
To prevent temporal leakage, for each forecasting anchor $t_a$ we select only the latest annual embedding slice strictly preceding the anchor,
$\tau_{y^\ast}<t_a$, and use
$\boldsymbol{z}_{y^\ast}(\boldsymbol{s})$
throughout both the training window and the forecast horizon.
Thus, AE enters the model as static exogenous spatial context available at forecast time.
\subsection{Implementation Details}

The latent spatial and temporal fields $f_s$ and $f_t$ in
Equations~\eqref{eq:baseline} and~\eqref{eq:baseline+AE}
follow the VAE-based parameterization of \citet{manring2025bstpp}.
For both the event-only and AE-augmented arms, the GP prior is reparameterized with PriorVAE. Spatial quadrature uses a $25\times25$ grid, and time is discretized into $n_t=50$ uniform bins over $[t_1,t_2)$. We fit each model with SVI for 4{,}000 steps at learning rate $10^{-2}$, and draw 1{,}000 posterior samples from the variational guide for evaluation.

Held-out predictive log densities on
$R_{\text{out}}\times[t_1,t_2)$
are computed by log-mean-exp aggregation across posterior samples.
Temporal integrals use fractional bin overlaps with $[t_1,t_2)$, while spatial integrals use polygon overlays between the computational grid and $R_{\text{out}}$ with exact area fractions.

\subsection{History--Context Trade-off}
\begin{figure}[t]
    \centering
    \begin{subfigure}[t]{0.49\linewidth}
        \centering
        \includegraphics[width=\linewidth]{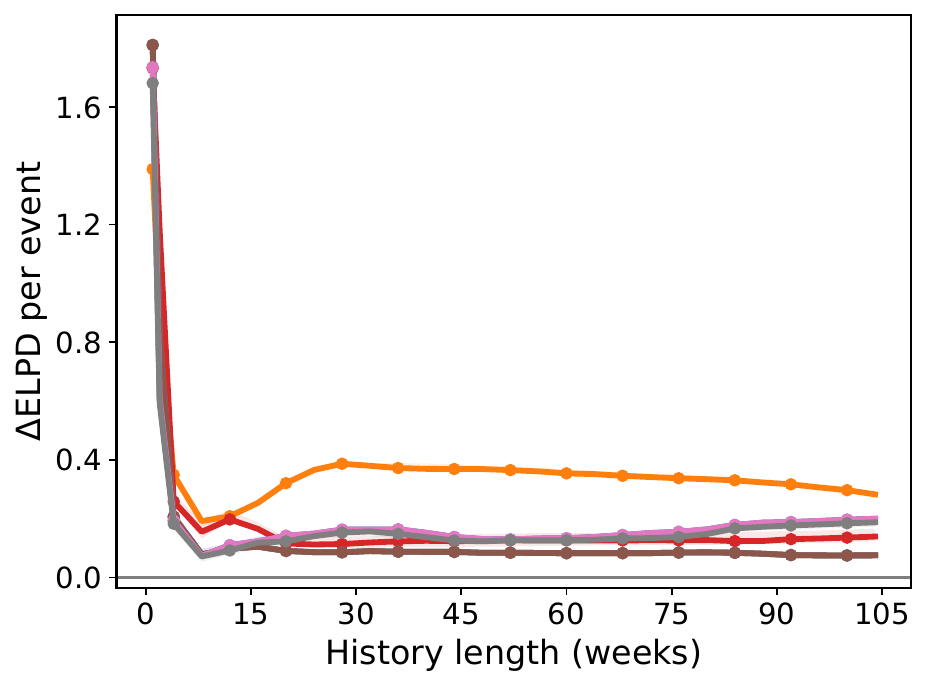}
    \end{subfigure}
    \hfill
    \begin{subfigure}[t]{0.49\linewidth}
        \centering
        \includegraphics[width=\linewidth]{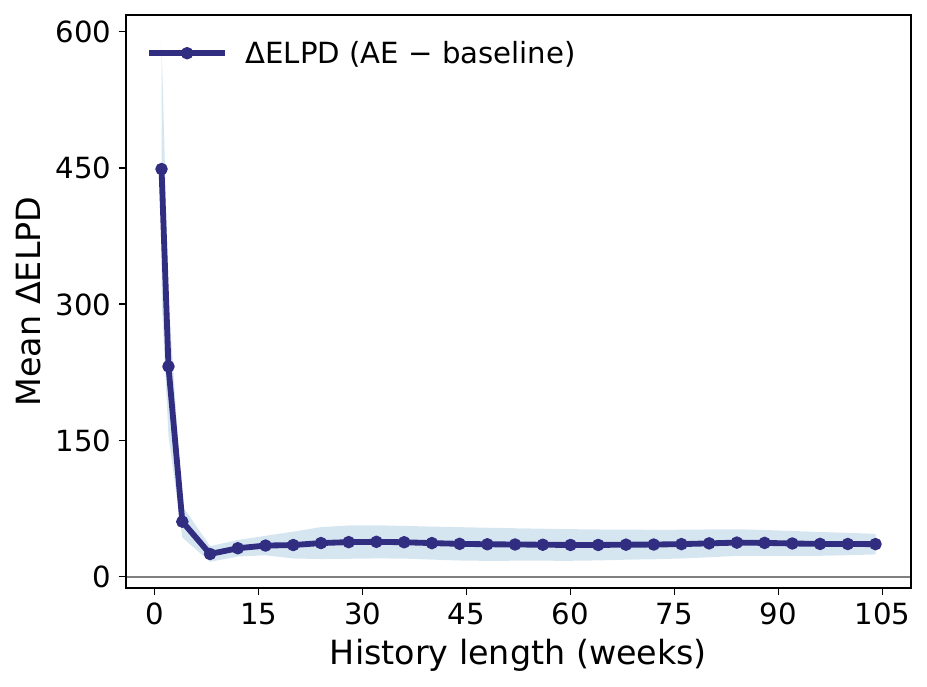}
    \end{subfigure}
    \caption{
History--context trade-off across held-out spatial regions.
\textbf{Left:} Per-mask paired per-event log-score difference
$\Delta_e(w;m)$ versus history length $w$.
\textbf{Right:} Aggregate trend across masks with normal-approximate uncertainty bands.
Values above $y=0$ favour the AE-augmented model.
}

    \label{fig:delta-history-context}
\end{figure}

Figure~\ref{fig:delta-history-context} summarizes how the predictive value of AE
changes with the available event-history length $w$.
The left panel shows the paired per-event log-score difference
$\Delta_e(w;m)$ for each held-out spatial mask, aggregated across forecast
anchors. The right panel reports the corresponding aggregate trend across
masks. Across the entire history range, the curves remain above the parity
line, indicating that the AE-augmented model consistently improves
out-of-region predictive performance relative to the event-only baseline.

\begin{figure}[t]
    \centering
    \begin{subfigure}[t]{0.49\linewidth}
        \includegraphics[width=\linewidth]{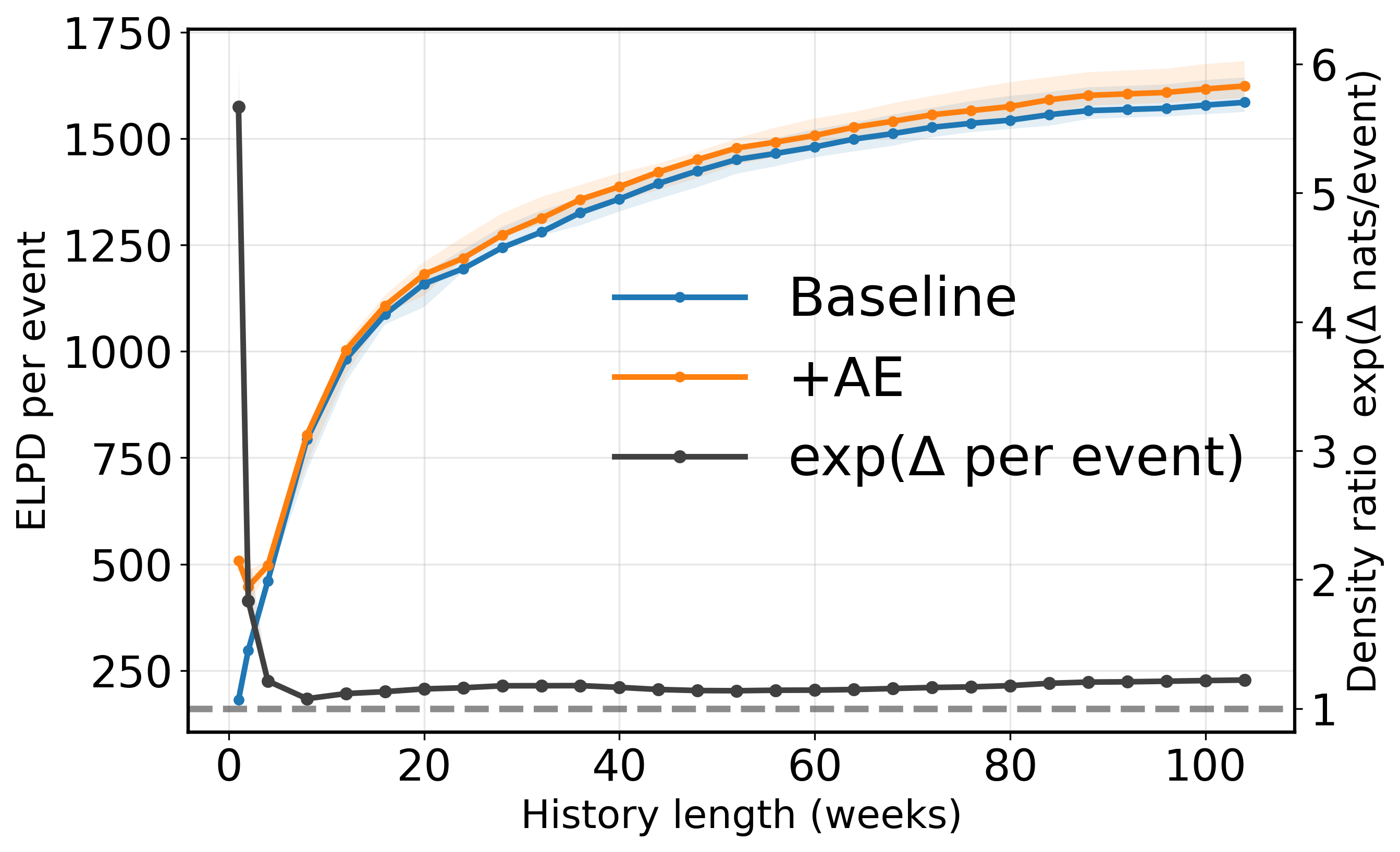}
    \end{subfigure}
    \hfill
    \begin{subfigure}[t]{0.49\linewidth}
        \includegraphics[width=\linewidth]{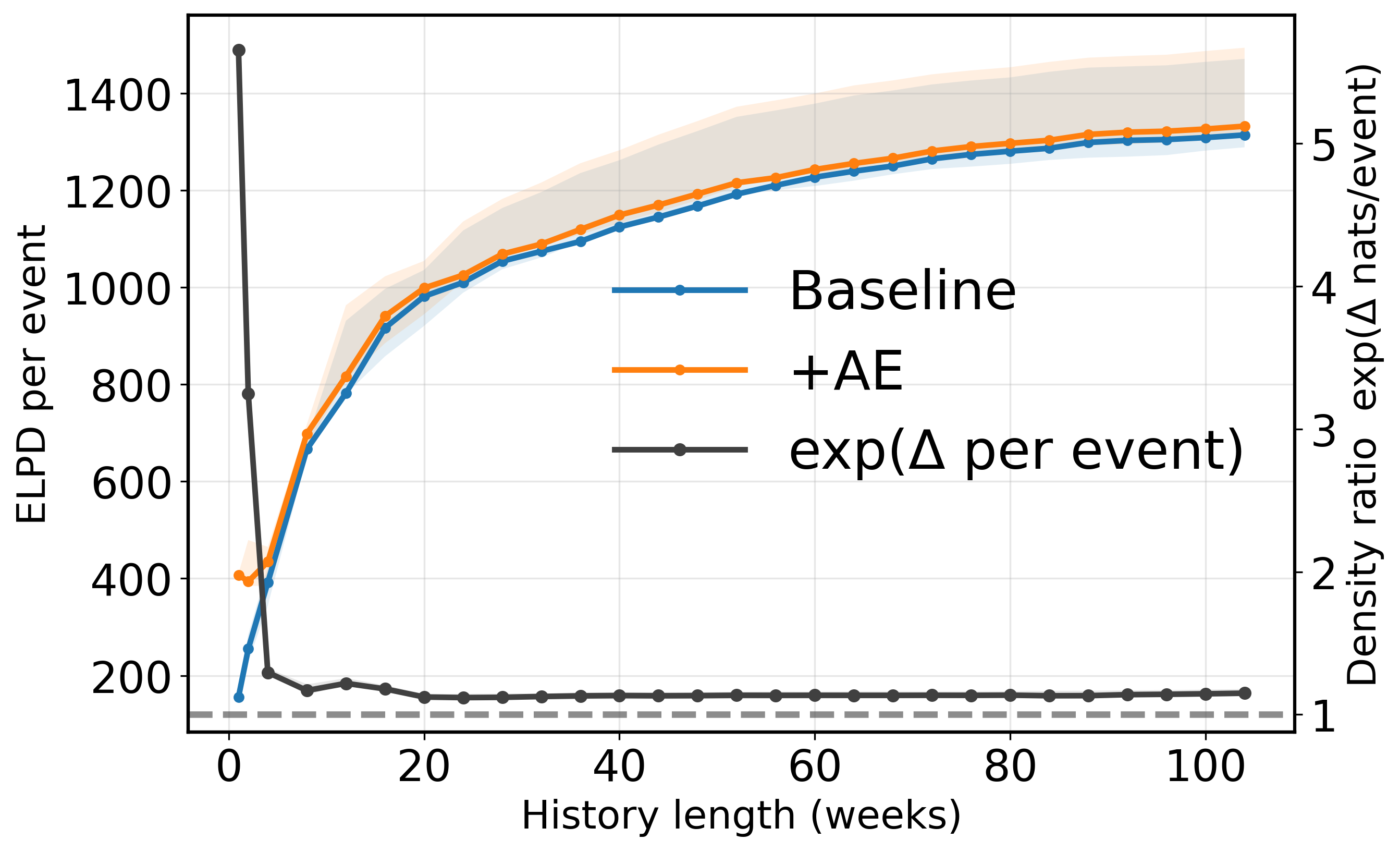}
    \end{subfigure}
    \caption{
    Held-out ELPD curves for two representative spatial masks.
    Blue: event-only baseline. Orange: AE-augmented model.
    Black: density ratio $\exp\{\Delta_e(w;m)\}$, with parity at $y=1$.
    AE gives the largest advantage in sparse-history regimes.
    }
    \label{fig:elpd-side-by-side}
\end{figure}

The gain is largest in sparse-history regimes and decreases as more events become available. At one to two weeks of training history, AE produces substantial improvements in predictive density, corresponding to roughly $2$--$6\times$ multiplicative gains over the event-only baseline. The advantage narrows as $w$ increases, but remains positive through long-history regimes, including $20$--$104$ weeks of training data. This pattern supports the central hypothesis of the paper: exogenous spatial context is most valuable when endogenous event evidence is weak, while continuing to complement event history even after substantial data accumulate. Minor non-monotonicity across weeks reflects the fact that longer histories add older events that may be less aligned with a given forecast anchor, producing finite mask-anchor variability around the dominant trend. Figure~\ref{fig:elpd-side-by-side} shows this pattern directly on total held-out ELPD for two representative masks. In both regions, the AE-augmented model achieves higher held-out ELPD than the event-only baseline for every history length. The separation is largest at short histories and narrows as more data are included. The secondary axis reports $\exp\{\Delta_e(w;m)\}$, the geometric-mean predictive-density ratio, which makes the scale of the improvement directly interpretable.

\begin{figure}[t]
    \centering
    \includegraphics[width=\linewidth]{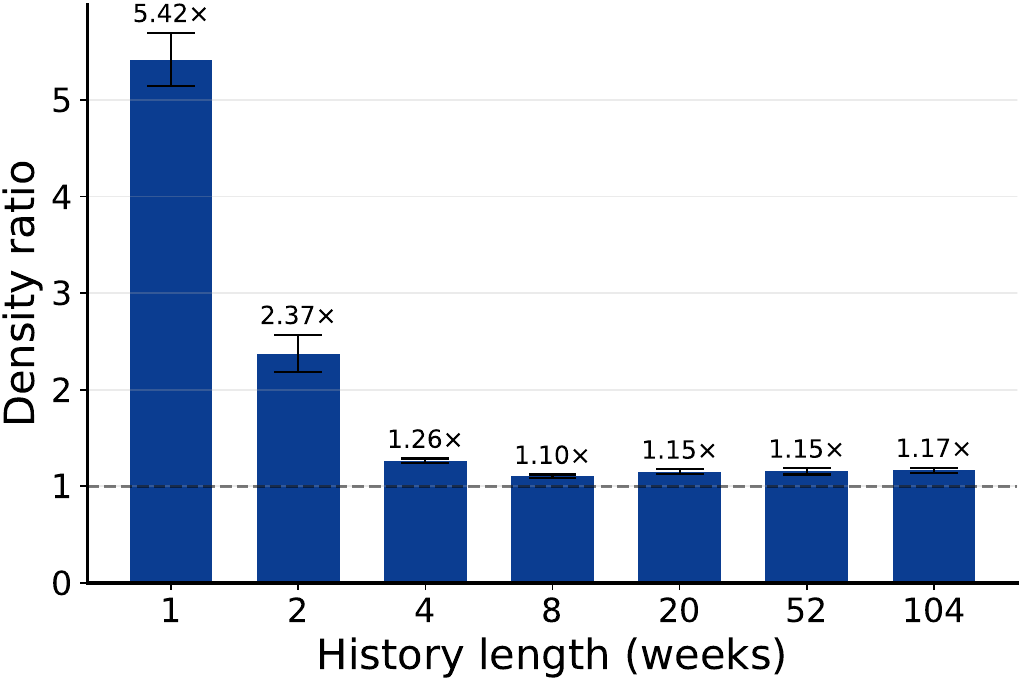}
    \caption{
    Mean multiplicative improvement in held-out predictive density versus history length. Bars report $\exp\{\Delta_e(w;m)\}$ averaged across masks, with normal-approximate descriptive intervals. Values above the parity line $y=1$ indicate higher predictive density under the AE-augmented model.}
    \label{fig:bars}
\end{figure}

\begin{figure}[t]
    \centering
    \includegraphics[width=\linewidth]{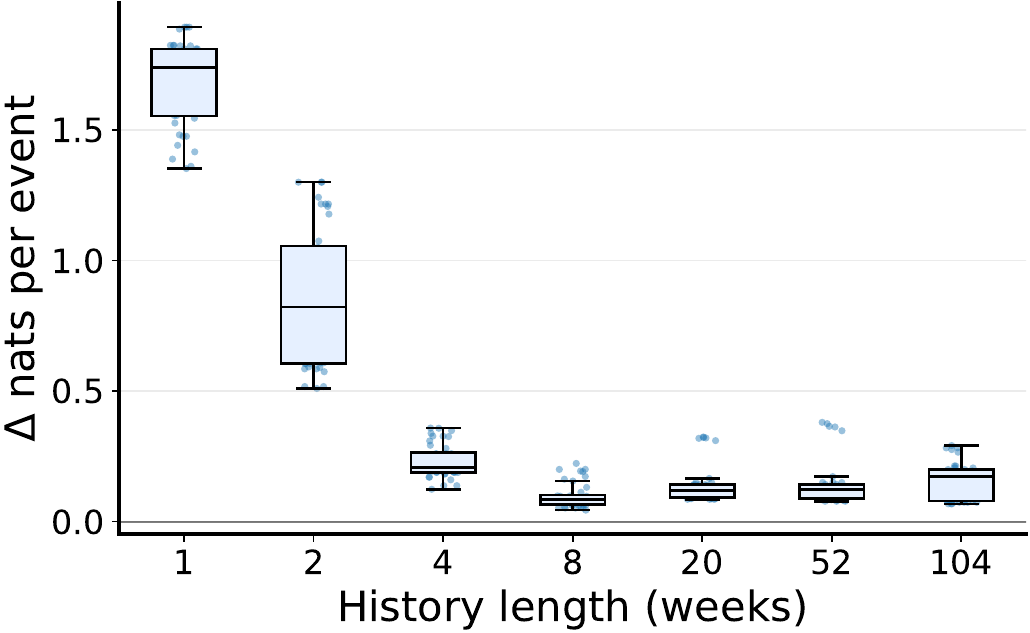}
    \caption{
    Distribution of paired per-event log-score improvements $\Delta_e(w;m)$ across masks at selected history lengths. Each point represents the median improvement for one mask across anchors; boxes show the interquartile range and median. Improvements remain above the parity line $y=0$ for all selected history lengths, with the largest gains at $w\in\{1,2\}$ weeks.
    }
    \label{fig:box_swarm}
\end{figure}

Figure~\ref{fig:bars} summarizes the same effect as a multiplicative
predictive-density ratio. Values above one indicate that AE assigns higher
predictive density to the realized held-out events than the event-only model.
The mean density ratio is $5.42\times$ at one week and $2.37\times$ at two
weeks, then decreases while remaining above parity for all selected history
lengths. Figure~\ref{fig:box_swarm} confirms that this effect is not driven by
a small subset of masks: the distribution of paired per-event log-score
differences remains strictly positive across regions, with the largest gains
concentrated in the shortest-history regimes.

\subsection{Posterior Field Structure}

\begin{figure}[t]
    \centering
    \captionsetup[subfigure]{labelformat=empty}

    \begin{subfigure}[t]{0.48\linewidth}
        \includegraphics[width=\linewidth]{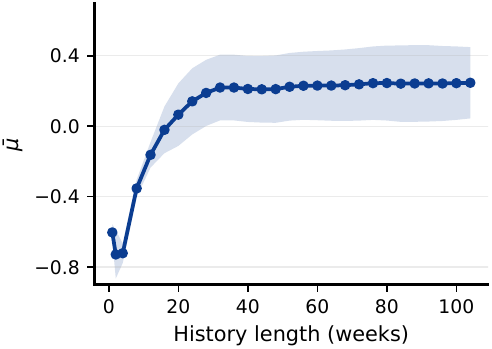}
    \end{subfigure}
    \hfill
    \begin{subfigure}[t]{0.48\linewidth}
        \includegraphics[width=\linewidth]{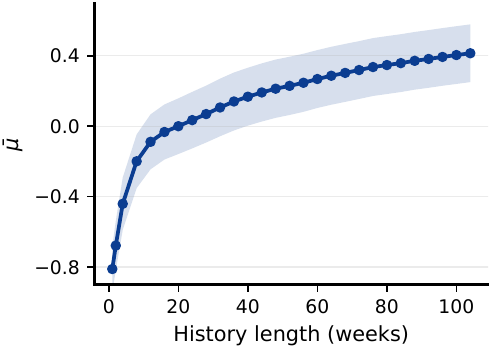}
    \end{subfigure}

    \vspace{3pt}

    \begin{subfigure}[t]{0.48\linewidth}
        \includegraphics[width=\linewidth]{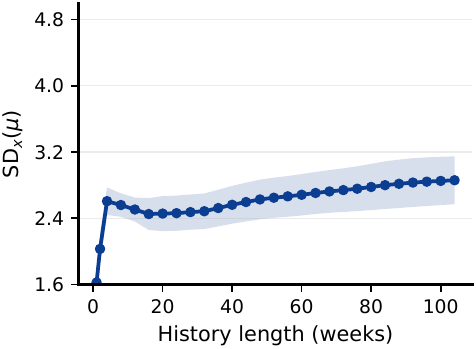}
    \end{subfigure}
    \hfill
    \begin{subfigure}[t]{0.48\linewidth}
        \includegraphics[width=\linewidth]{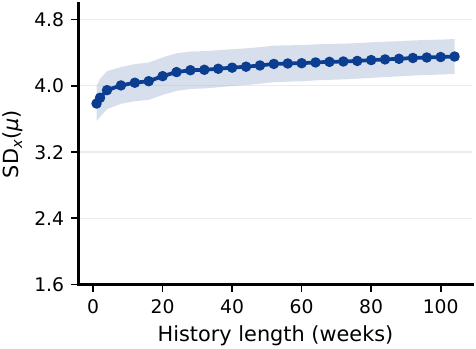}
    \end{subfigure}
    \caption{
    Posterior spatial-field summaries versus history length. Top: spatial mean log-intensity $\bar{\mu}(w)$. Bottom: spatial contrast $\mathrm{SD}_x(\mu;w)$. Left: AE-augmented model; right: event-only baseline. Bands summarize across-mask variability.
    }
    \label{fig:mecanistic}
\end{figure}

The predictive gains are accompanied by systematic differences in the fitted
latent spatial fields. Figure~\ref{fig:mecanistic} summarizes two posterior
field statistics as the history length $w$ increases.
The top row reports $\bar{\mu}(w)$, the spatial average of the posterior mean
log-intensity over the held-out region.
Both models rapidly adjust their overall intensity level as history becomes
available and then plateau, but the AE-augmented model converges more quickly
in the low-history regime.

The bottom row reports $\mathrm{SD}_x(\mu;w)$, the spatial standard deviation
of posterior mean log-intensity across the held-out region.
The AE-augmented model produces consistently smoother spatial fields than the
event-only baseline, while both models develop stronger spatial contrast as
more history accumulates.
Taken together, these summaries suggest that AE improves sparse-history
forecasting by stabilizing early estimates of the held-out risk surface, while
its smaller long-history gains reflect continued refinement of spatial
structure rather than correction of gross level errors.

\section{RELATED WORK} 
Classical STPPs provide the probabilistic foundation for modeling events in continuous space and time. Frameworks such as IPPs provide interpretable risk maps by linking the log-intensity to observed features via likelihood-based inference \citep{gonzalez2016spatio,moraga2023spatial}. LGCPs extend IPPs by imposing a latent Gaussian field over the log-intensity, capturing residual clustering and non-stationarity beyond measured covariates \citep{moller1998log}. Recent toolchains have improved the scalability and routine use of LGCPs for both point-level and aggregated outcomes \citep{watson2024twenty}. These models are widely used in epidemiology for spatial smoothing and uncertainty quantification \citep{amaral2023spatio,meyer2017spatio}.
In recent years, deep learning has emerged as a powerful paradigm for STPPs, demonstrating superior predictive performance and the ability to capture highly complex dependencies. This line of work includes models based on neural ODEs \citep{chen2020neural} and, more recently, score-based generative models that frame event generation as a diffusion-denoising process \citep{yuan2023spatio,ludke2024unlocking}. The flexibility of these architectures comes with reduced interpretability and higher computational demands, which can limit adoption in high-stakes settings.

Integrating exogenous covariates is key to improving both explanation and prediction. Best practices in spatial statistics advocate a systematic covariate workflow that includes exploratory analysis, mechanistic justification, regularization, and rigorous out-of-sample evaluation \citep{giorgi2021model}. Within the Cox-process family, covariates generally influence the mean structure of the latent field, which in turn accounts for unobserved heterogeneity \citep{moller1998log,watson2024twenty}. Toolkits designed for environmental health and epidemiology offer practical approaches for effective integration \citep{moraga2023spatial,meyer2017spatio}. On the temporal side, \citep{meng2024transfeat} proposes an interpretable transformer that explicitly disentangles the influence of event history from covariate representations, enabling attribution of covariate importance without sacrificing predictive accuracy. 

EMS demand modeling has long treated calls as inhomogeneous Poisson events on fine space–time grids \citep{zhou2015spatio,zhou2016predictingg}. To address sparsity and periodicity, time-varying Gaussian mixtures fix spatial components while allowing mixture weights to evolve, improving accuracy over operational heuristics \citep{zhou2015spatio}. Nonparametric baselines reweight historical incidents by recency and relevance using spatio-temporal kernel–density estimation \citep{zhou2015predicting}, and kernel warping adapts to complex urban geometry by transporting kernels to a historical point cloud \citep{zhou2016predicting}. Because interventions occur on road networks, linear-network STPPs with nonseparable space–time interactions sharpen fit and localization \citep{gilardi2024nonseparable}. Complementary modeling choices capture additional data characteristics: zero-inflated Poisson regression improves calibration under structural zeros \citep{steins2019forecasting}, and self-exciting formulations capture short-term clustering \citep{li2019nonparametric}. 

\section{CONCLUSION}

\paragraph{Main finding.} 
{This work examined when exogenous spatial context can complement limited event history in spatio-temporal point-process forecasting.} Using matched LGCP arms, we isolated the contribution of AlphaEarth embeddings under spatial hold-out evaluation. Across eight held-out regions in Montgomery County, Pennsylvania, the AE-augmented model consistently outperformed the event-only baseline. The gains were largest with short histories, reaching $2$ to $6\times$ multiplicative improvements with only one to two weeks of training events, and remained positive with longer histories. This suggests that contextual information can stabilise spatial transfer when local event evidence is weak. {The gains are plausibly driven by persistent properties of place that AE embeddings may encode, such as land cover, built form, settlement density, infrastructure, and other remote-sensing-derived proxies. We do not interpret individual AE dimensions causally. Instead, AE is treated as a compact external representation of spatial context whose predictive value is evaluated under a controlled spatial-transfer protocol.}

\paragraph{Implications for STPPs.}
More broadly, the results suggest that context should be treated as a core modelling component rather than an optional extension. Spatial and temporal domains are increasingly accompanied by structured external information, including remote-sensing embeddings, mobility fields, land-use layers, weather, infrastructure, and other contextual signals. Event histories may recover some of these effects indirectly, but only after enough events have been observed. This motivates more systematic use of contextual information in flexible neural and representation-rich STPPs.

\paragraph{Operational relevance.}
For EMS forecasting, context appears most useful in operationally difficult regimes, such as newly developed areas, boundary regions, or locations with limited recent call history. Better spatial transfer in these settings could support more reliable demand estimation and preparedness. At the same time, probabilistic gains are not deployment value by themselves. Future work should connect these improvements to decision-facing evaluations, including response time, unit availability, and other operational constraints. 


\paragraph{Limitations and Future Directions.}
Our evaluation focuses on proper predictive scores---held-out ELPD and per-event log-density contrasts---rather than operational outcomes. A natural next step is to assess decision-centric metrics such as Predictive Accuracy Index, calibration diagnostics, and CRPS, and to connect forecast quality to downstream EMS decisions through simulation-based evaluation. Methodologically, the present study establishes the value of contextual information within a transparent LGCP backbone. The next step is to develop STPP architectures that integrate high-dimensional contextual fields more flexibly, including neural models, while preserving reliable spatial generalisation. In parallel, the field needs importance measures tailored to contextual information in point-process models, so that improved forecasts can also yield sharper understanding of the signals driving spatio-temporal risk.

\bibliographystyle{named}
\bibliography{References}

\end{document}